\documentclass[runningheads]{llncs}
\usepackage[T1]{fontenc}
\usepackage{graphicx}
\usepackage{amsmath}
\usepackage{algorithm}
\usepackage{algorithmic}
\usepackage{amsfonts}
\usepackage{listings}
\usepackage{xcolor}
\usepackage{hyperref} 
\usepackage{orcidlink}

\definecolor{codegray}{gray}{0.9}
\definecolor{keywordcolor}{RGB}{255, 0, 0}
\definecolor{commentcolor}{RGB}{0, 128, 0}
\definecolor{stringcolor}{RGB}{0, 0, 255}

\begin{document}
\renewcommand{\thefootnote}{\fnsymbol{footnote}}  
\title{UMambaAdj: Advancing GTV Segmentation for Head and Neck Cancer in MRI-Guided RT with UMamba and nnU-Net ResEnc Planner}
\titlerunning{UMambaAdj for Head and Neck Cancer Tumor Segmentation}
%
\author{Jintao Ren \thanks{Corresponding author: jintaoren@clin.au.dk}\inst{1,2}\orcidID{0000-0002-1558-7196} \and 
Kim Hochreuter \inst{1,2}\orcidID{0009-0000-7347-1688} \and 
Jesper Folsted Kallehauge \inst{1,2}\orcidID{0000-0003-3705-5390} \and 
Stine Sofia Korreman\inst{1,2,3}\orcidID{0000-0002-3523-382X} }
\authorrunning{J. Ren et al.}
%
\institute{Aarhus University, Department of Clinical Medicine, Nordre Palle Juul-Jensens Blvd. 11,
8200 Aarhus, Denmark \and
Aarhus University Hospital, Danish Centre for Particle Therapy, Palle Juul-Jensens Blvd. 25, 8200 Aarhus, Denmark \and
Aarhus University, Department of Oncology, Palle Juul-Jensens Blvd. 35,
8200 Aarhus, Denmark \\
\email{jintaoren@clin.au.dk}
\\ 
}
\maketitle              
\begin{abstract}
    Magnetic Resonance Imaging (MRI) plays a crucial role in MRI-guided adaptive radiotherapy for head and neck cancer (HNC) due to its superior soft-tissue contrast. However, accurately segmenting the gross tumor volume (GTV), which includes both the primary tumor (GTVp) and lymph nodes (GTVn), remains challenging. Recently, two deep learning segmentation innovations have shown great promise: UMamba, which effectively captures long-range dependencies, and the nnU-Net Residual Encoder (ResEnc), which enhances feature extraction through multistage residual blocks. In this study, we integrate these strengths into a novel approach, termed 'UMambaAdj'. Our proposed method was evaluated on the HNTS-MRG 2024 challenge test set using pre-RT T2-weighted MRI images, achieving an aggregated Dice Similarity Coefficient (DSC$_{agg}$) of 0.751 for GTVp and 0.842 for GTVn, with a mean DSC$_{agg}$of 0.796. This approach demonstrates potential for more precise tumor delineation in MRI-guided adaptive radiotherapy, ultimately improving treatment outcomes for HNC patients. Team: DCPT-Stine's group.

\end{abstract}
\section{Introduction}
Magnetic Resonance Imaging (MRI) plays a pivotal role in radiotherapy (RT), particularly in MRI-guided adaptive radiotherapy, due to its superior soft-tissue contrast compared to other imaging modalities like computed tomography (CT). This soft-tissue contrast enables more accurate delineation, which is especially crucial in head and neck cancer (HNC), where the intricate anatomy and proximity of vital structures, such as the salivary glands, optic nerves, and spinal cord \cite{ahmed2010value,brouwer2015ct}, make precise tumor targeting critical. MRI's ability to differentiate between tumor tissues and surrounding normal tissues enhances radiation delivery accuracy, reducing the risk of collateral damage to critical structures \cite{benitez2024mri,mohamed2018prospective,mcdonald2024use}.

Despite these advantages, accurate delineation of HNC tumors, including both the primary tumor volume (GTVp) and involved nodal metastasis (GTVn), remains challenging. The heterogeneous and diffuse nature of HNC tumors often makes obtaining clear margins difficult \cite{Ruhle2023,jensen2020danish}. Additionally, MRI's lower spatial resolution in the third dimension (through-slice direction) compared to the in-plane resolution can complicate tumor delineation, leading to variability in interpretation among clinicians and resulting in significant inter-observer variation (IOV).

Given these challenges, there is a growing need for automated, accurate segmentation methods to enhance the consistency and precision of tumor delineation in MRI-guided adaptive radiotherapy. Deep learning-based medical image segmentation has emerged as a promising solution, often building on the classic U-Net architecture, known for its symmetrical encoder-decoder design and skip connections \cite{ronneberger2015u}.  It plays a crucial role in medical image analysis by identifying and delineating structures such as organs, lesions, tumors, and tissues across various 2D and 3D imaging modalities, including CT, PET, and MRI, thereby aiding in diagnosis, treatment planning, and prognoses.

Recently, the leading deep learning models for segmentation have shifted between convolutional neural networks (CNNs) and transformer-based architectures. CNNs excel at capturing translational invariances and local features but often face challenges with long-range dependencies \cite{luo2016understanding}. In contrast, Vision Transformers (ViTs) \cite{dosovitskiy2020image} effectively capture global context by treating the image as a sequence of patches. However, their self-attention mechanism incurs a quadratic computational cost relative to the number of patches \cite{gu2023mamba}, and Transformers tend to be prone to overfitting, especially when working with limited datasets \cite{lin2022survey,heidari2024computation}. Leveraging the complementary strengths of both architectures, many studies have explored hybrid models that integrate ViTs with CNNs, resulting in architectures such as nnFormer \cite{zhou2021nnformer}, TransUNet \cite{chen2021transunet}, UNETR \cite{hatamizadeh2022unetr}, SwinUNETR \cite{hatamizadeh2021swin}, and UNETR++ \cite{shaker2024unetr++}. These hybrid models have also gained popularity in HNC GTV segmentation. For instance, Hung Chu et al. \cite{chu2022swin} demonstrated that the SwinUNETR achieved an average Dice similarity coefficient (DSC) of 0.626 on CT/PET data, while a cross-modal Swin transformer achieved a mean DSC of 0.769 for GTVp using CT/PET modalities \cite{li2024swincross}. Despite these advancements, the U-Net architecture continues to be a foundational design in all segmentation models.

The field is now advancing with structured state-space models (SSMs), such as Mamba \cite{gu2021efficiently,gu2023mamba,dao2024transformers}, which offer improved segmentation performance by efficiently modeling long-range dependencies and scaling effectively with sequence length \cite{2024visual_mamba,heidari2024computation}. Mamba's ability to capture complex anatomical relationships makes it well-suited for segmenting GTVp and GTVn in HNC, as their locations are often closely correlated with each other. However, the optimal model configuration depends on task-specific factors such as the foreground-to-background ratio, image resolution, and tumor size variability, as each architecture's effectiveness varies with different imaging challenges and anatomical complexities.

Recently, two innovative approaches, UMamba \cite{ma2024u} and the new nnU-Net Residual Encoder (ResEnc) planner \cite{isensee2024nnu}, have gained significant attention in medical image segmentation. The default UMamba encoder incorporates a Mamba layer after each CNN block, which can be computationally expensive, especially at the first level where image features have a large resolution. This design, which includes both a residual encoder and decoder, can be cumbersome to train and provides only limited accuracy improvements in its default configuration \cite{isensee2024nnu}. In contrast, the nnU-Net ResEnc enhances feature extraction through multiple blocks of residual CNN encoding and employs only a single CNN layer in the decoder, offering a more efficient solution.

This study focuses on addressing the first task of the HNTS-MRG 2024 challenge, which aims to segment both GTVp and GTVn using pre-RT T2-weighted MRI images. We aim to improve gross tumor volume (GTV) segmentation in T2-weighted MRI for head and neck cancer by integrating the strengths of UMamba and nnU-Net ResEnc. We refer to this integrated approach as \textbf{UMambaAdj} in this study.

Our contributions are as follows:
\begin{itemize}
    \item We optimize UMamba by removing the Mamba layer in the first stage and the residual blocks in the decoder, significantly enhancing computational efficiency while preserving its ability to capture long-range dependencies in deeper stages.
    \item We combine UMamba’s long-range dependency modeling with nnU-Net ResEnc’s enhanced residual encoding to improve the accuracy of GTV delineation in the complex anatomy of head and neck cancer.
\end{itemize}

\section{Material and Methods}
\subsection{Data}
The dataset used in this study was provided by the organizers of the HNTS-MRG 2024 challenge task 1, consisting of 150 HNC patients, primarily with oropharyngeal cancer (OPC). Each patient had T2-weighted MRI sequences of the head and neck region, acquired at University of Texas MD Anderson Cancer Center \cite{wahid2024training}. The images included pre-RT scans taken 1-3 weeks before the start of radiotherapy. For all cases, GTV for the primary tumor (GTVp) and involved lymph nodes (GTVn) were independently segmented by 3 to 4 expert physician observers based on the MRI images. The ground truth segmentation was then generated using the Simultaneous Truth And Performance Level Estimation (STAPLE) algorithm.

\subsection{Network Architecture}
\begin{figure} 
    \includegraphics[width=\textwidth]{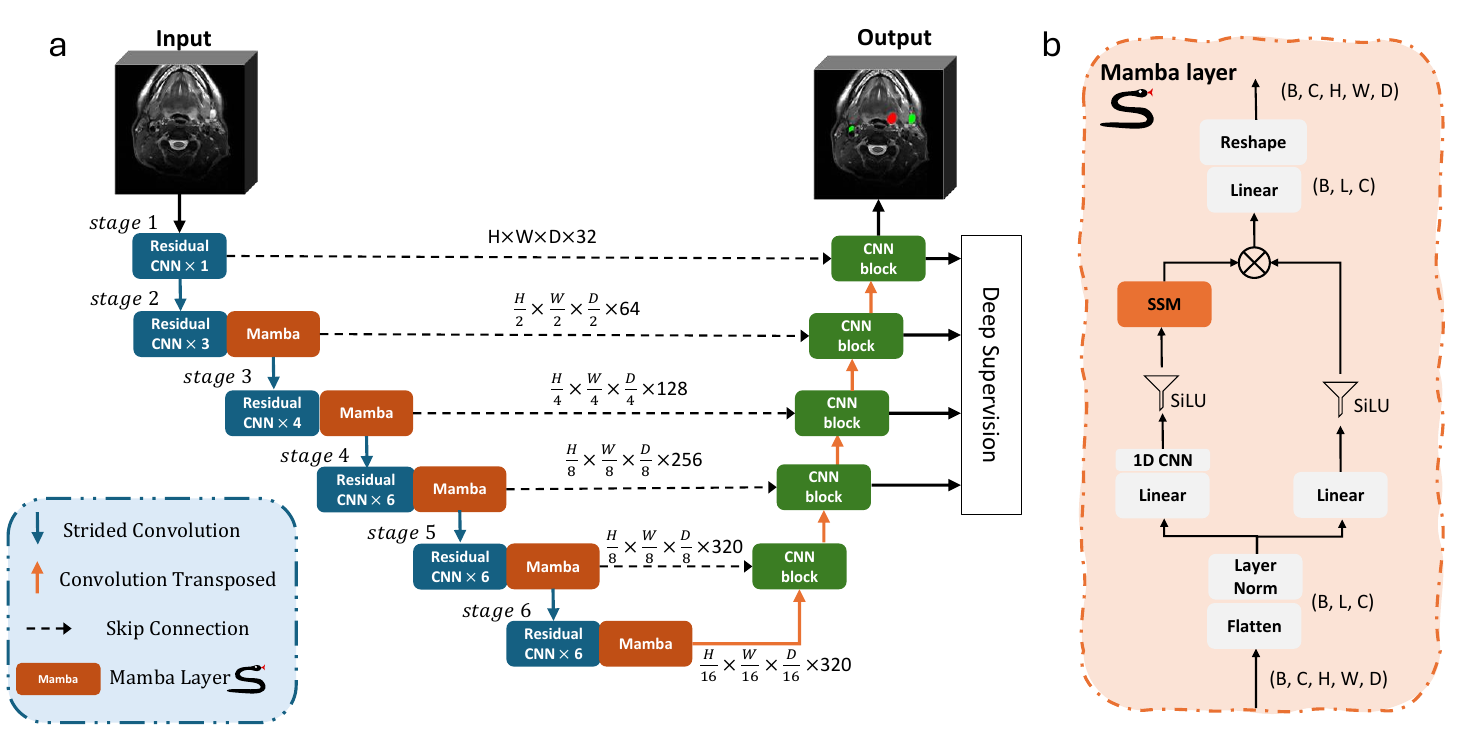}
    \caption{(a) Overview of the proposed UMamba adjustment (UMambaAdj) network architecture. (b) Details of the Mamba layer.}\label{figure1}
    \end{figure}

The proposed network architecture is based on a combination of a 3D ResEnc U-Net and Mamba blocks. The CNN part of  network architecture was designed according to the new nnU-Net Residual encoder planner (M). The U-Net consists of 6 stages, each with varying features per stage (32, 64, 128, 256, 320, 320). The network uses 3D convolutional layers with kernel sizes mostly set to (3, 3, 3), except for the first stage where it is (1, 3, 3). The strides vary across stages to enable down-sampling at different levels, with a stride set of (1, 2, 2) between the first and second stages, and (2, 2, 2) for the remaining stages. Each stage contains a different number of residual CNN blocks with counts of (1, 3, 4, 6, 6, 6) in the encoder, and a Mamba layer is appended after each residual CNN block except the first stage. A skip connection with concatenation was connecting the Mamba layer and the decoder blocks, while each decoder block consists of only one 3D CNN block. Instance normalization and the Leaky ReLU activation function are used. Additionally, deep supervision is applied at the top four levels of the network outputs. The overall structure of the network can be seen in \autoref{figure1}a. 

\subsection{Mamba layer}
The Mamba layer, adapted from the UMamba design for capturing long-range dependencies, processes input image feature maps of shape $(B, C, H, W, D)$, where $B$ is the batch size, $C$ the channel, and $H, W, D$ the spatial dimensions. These feature maps are first reshaped and transposed into a flattened representation $(B, L, C)$, where $L = H \times W \times D$, treating all spatial locations as individual patch tokens. The reshaped features are then normalized using Layer Normalization. Following normalization, the features undergo two parallel branches. In the first branch, the features are expanded to $(B, 2L, C)$ through a linear transformation, followed by a 1D convolutional layer and a SiLU activation function, ultimately passing through the SSM layer. In the second branch, a similar expansion process is performed via a linear transformation and a SiLU activation function, but without the convolutional or SSM steps. The outputs from the two branches are then combined using element-wise (Hadamard) multiplication. The resulting features are projected back to the original token dimension, reshaped, and transposed to restore the original input shape of $(B, C, H, W, D)$, maintaining the spatial structure of the image feature maps for further processing. The detailed flow of a Mamba layer can be seen in \autoref{figure1}b.

\subsection{Training Parameters}
Training was conducted with a batch size of 4, using a patch size of (48, 192, 192) and Z-score normalization for data preprocessing. The median image size in voxels was (123, 512, 511), with spacing set at (1.199, 0.5, 0.5). Resampling spline interpolation functions were employed to adjust both image and segmentation data, using an interpolation order of 3 for images and an order of 1 (linear) for masks. The training was performed using the SGD optimizer with a PolyLR scheduler (exponent = 0.9), starting with a learning rate of 0.01. The adoption of the Mamba layer often led to gradient vanishing or explosion during training, especially when using mixed-precision (fp16) with automatic casting. To address this, normalized gradients were clipped with a value of 1.
    
The 150 patients were randomly divided into 5 folds, with each fold comprising 120 patients for training and 30 for validation. Each model was trained for a maximum of 1,000 epochs, and the final models from the last epoch were saved for prediction. For the final challenge submission, predictions on the test set were generated using an ensemble of all models trained across the five folds. In line with reproducibility and verification guidelines \cite{hurkmans2024joint}, all source code, predicted masks, training logs and trained weights have been made publicly available on GitHub\footnote{\url{https://github.com/Aarhus-RadOnc-AI/UMambaAdj}}.
    
\subsection{Evaluation}
The aggregated Dice Similarity Coefficient (DSC$_{agg}$) \cite{andrearczyk2021overview} was used as the primary evaluation metric in accordance with the guidelines of the HNTS-MRG 2024 challenge. Additionally, we employed the mean 95th percentile Hausdorff Distance (HD$_{95}$), and the mean surface distance (MSD) as supplementary metrics to further evaluate the segmentation performance for both GTVp and GTVn. Hausdorff Distance (HD) was used for case study. 

To evaluate the performance of the proposed method, we compared both the segmentation accuracy and training epoch times of the default nnU-Net, nnU-Net Residual Encoder (ResEnc), UMamba Encoder (UMambaEnc), and the proposed UMambaAdj using the DSC$_{agg}$ metric. Since the Mamba block involves multiple tensor shape manipulations that are not fully represented by FLOPs, we measured the stable epoch time after the first epoch as a direct indicator of model efficiency. All compared groups were trained with a batch size of 4. 

\subsection{System environment}
The experiments were conducted on a system equipped with dual AMD Ryzen Threadripper 3990X 64-core processors (128 threads) and 256GB of system memory. An NVIDIA RTX A6000 GPU with 48GB VRAM was used for training. The software environment included Python 3.12.4, PyTorch 2.4.0, CUDA 12.6 and nnU-Net 2.5.1. Distance metrics were calculated using SimpleITK 2.4.0.

\section{Results}
\subsection{Cross-Validation performance}
\begin{table}[h!]
    \centering
    \caption{GTVp performance on 5-Fold cross-validation}\label{table1}
    \begin{tabular}{lcccccc}
    \hline
    \textbf{Metric} & \textbf{Fold 0} & \textbf{Fold 1} & \textbf{Fold 2} & \textbf{Fold 3} & \textbf{Fold 4} & \textbf{Average} \\
    \hline
    \textbf{DSC$_{agg}$} & 0.804 & 0.742 & 0.804 & 0.774 & 0.776 & 0.779 \\ 
    \textbf{HD$_{95}$ [mm]}   & 5.7   & 8.1  & 6.8   & 6.5  & 10.4  & 7.5   \\ 
    \textbf{MSD [mm]}    & 2.6   & 2.8   & 1.9   & 1.9   & 4.2   & 2.68   \\
    \hline
    \end{tabular}
    \end{table}
    \vspace{-20pt} 

    \begin{table}[h!]
    \centering
    \caption{GTVn performance on 5-Fold cross-validation}\label{table2}
    \begin{tabular}{lcccccc}
    \hline
    \textbf{Metric} & \textbf{Fold 0} & \textbf{Fold 1} & \textbf{Fold 2} & \textbf{Fold 3} & \textbf{Fold 4} & \textbf{Average} \\
    \hline
    \textbf{DSC$_{agg}$} & 0.874 & 0.849 & 0.751 & 0.875 & 0.885 & 0.847 \\
    \textbf{HD$_{95}$ [mm]}   & 15.2  & 15.4  & 24.5 & 21.2  & 17.6  & 18.78  \\
    \textbf{MSD [mm]}    & 3.1 & 2.6   & 4.2 & 3.6   & 3.3   & 3.36    \\
    \hline
    \end{tabular}
    \end{table}
The performance metrics for GTVp and GTVn across 5-fold cross-validation are summarized in \autoref{table1} and \autoref{table2}. For GTVp, DSC$_{agg}$ ranged from 0.742 to 0.804 across different folds, with an average of 0.779. The HD$_{95}$ varied from 5.7 mm to 10.4 mm, yielding an average of 7.5 mm. MSD ranged between 1.9 mm and 4.2 mm, with an average of 2.68 mm. For GTVn, DSC$_{agg}$ ranged from 0.751 to 0.885, with an average of 0.847. The HD$_{95}$ showed a wider range from 15.2 mm to 24.5  mm, resulting in an average of 18.78 mm. The MSD values spanned from 2.6 mm to 4.2 mm, with an average of 3.36 mm. 

\subsection{Comparision of nnU-Net, ResEnc, UmambaEnc and UmambaAdj}
\begin{table}    
    \centering
    \caption{Performance comparison between nn-UNet default, ResEnc, UMambaEnc, and the proposed UMambaAdj.}\label{table3}
    \begin{tabular}{l|rrr|rrr}
    \hline
    \multicolumn{1}{c|}{} & \multicolumn{3}{c|}{GTVp} & \multicolumn{3}{c}{GTVn} \\ \hline
    Methods               & \multicolumn{1}{l}{DSC$_{agg}$} & \multicolumn{1}{l}{HD$_{95}$} & \multicolumn{1}{l|}{MSD} & \multicolumn{1}{l}{DSC$_{agg}$} & \multicolumn{1}{l}{HD$_{95}$} & \multicolumn{1}{l}{MSD} \\ \hline
    nnUNet default        & 0.78                                & 14.5                              & 3.7                               & 0.859                               & 28.3                                & 4.4                             \\
    nnUNet ResEnc         & 0.803                               & 10.5                              & 5.5                               & 0.872                               & 15.8                              & 3.1                              \\
    UMambaEnc             & 0.794                               & 8.7                               & 3.5                               & \textbf{0.880}                      & \textbf{10.5}                      & \textbf{2.2}                     \\
    UMambaAdj             & \textbf{0.804}                      & \textbf{5.7}                      & \textbf{2.6}                      & 0.874                               & 15.2                              & 3.1                             \\ \hline
    \end{tabular}\\
    \footnotesize{\textbf{Bold} numbers indicate the best performance for each metric.}
\end{table}

\begin{figure}[h!]\centering
    \includegraphics[width=1\textwidth]{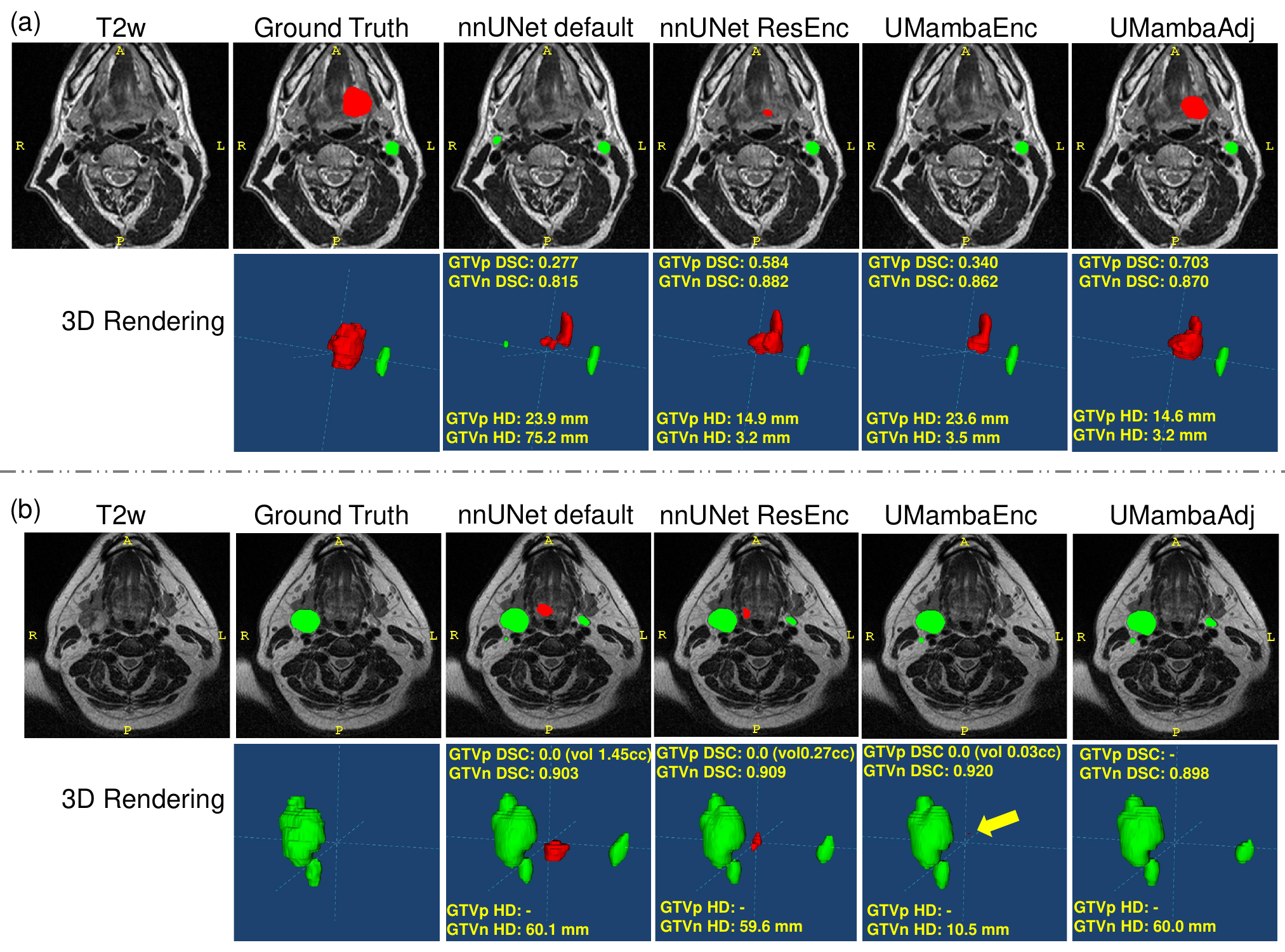}
    \caption{ Two patients (a and b) were selected for illustration. For each patient, the first row (left to right) displays the original T2-weighted MRI image, the ground truth overlaid on the image, and the segmentation results from all compared methods overlaid on the image. The second row shows the 3D renderings of the delineations and segmentations. \textcolor{red}{Red} represents the GTVp, and \textcolor{green}{Green} represents the GTVn. Segmentation metrics, including DSC and HD, are shown on the rendering subfigures. A yellow arrow in patient (b) indicates a nearly invisible small false-positive GTVp segmentation predicted by UMambaEnc.}\label{figure2}
    \end{figure}

\autoref{table3} summarizes the performance of various models for GTVp and GTVn segmentation. For GTVp, the proposed UMambaAdj achieved the highest DSC$_{agg}$ (0.804) and the best results in terms of HD$_{95}$ (5.7 mm) and MSD (2.6 mm). In the case of GTVn, UMambaEnc achieved the highest DSC$_{agg}$ (0.880) and outperformed others with the lowest HD$_{95}$ (10.5 mm) and MSD (2.2 mm).

Two cases were selected for illustration in \autoref{figure2}. For patient (a), all methods except UMambaAdj predicted a significantly smaller GTVp (DSC range 0.277–0.584), with the lower part missing, whereas UMambaAdj achieved a higher DSC of 0.703, despite all methods failing to capture the upper part. Additionally, the default nnU-Net model incorrectly identified a lymph node as GTVn, resulting in an HD of 75.2 mm, compared to 3.2–3.5 mm for the other methods. For patient (b), all methods except UMambaAdj made false positive predictions of GTVp in the same location. Moreover, all methods except UMambaEnc incorrectly predicted a lymph node as positive bilaterally, leading to an HD of 60 mm.

The training epoch time for the nnU-Net default model was 116 seconds, while the nnU-Net ResEnc took 127 seconds. The UMambaEnc required 400 seconds, and UMambaAdj took 199 seconds. Mixed-precision (fp16) autocast was enabled for all layers, except the Mamba layer, to ensure training stability.

\subsection{Final test score}
We submitted our trained UMambaAdj models in a Docker container to the HNTS-MRG 2024 challenge on the grand challenge platform. Predictions were made using an ensemble of all five  models trained across the 5-folds. Our model was evaluated on the test set using pre-RT T2-weighted MRI images, achieving an DSC$_{agg}$ of 0.751 for GTVp and 0.842 for GTVn, resulting in an overall mean DSC$_{agg}$ of \textbf{0.796}.

\section{Discussion}
In this study, we developed a customized network that integrates features from both UMamba and the nnU-Net Residual Encoder for T2-weighted MRI head and neck tumor segmentation. The aim was to combine the feature extraction strength of the residual encoder with the long-range dependency capabilities of Mamba blocks. Compared to the original UMambaEnc, the proposed UMambaAdj demonstrated comparable segmentation accuracy with reduced training and inference time, and outperformed UMambaEnc for GTVp. It also achieved significantly better HD$_{95}$ and MSD while maintaining similar DSC$_{agg}$ compared to nnU-Net ResEnc. All recent methods outperformed the default nnU-Net model across all metrics, confirming the complementary strengths of UMamba and nnU-Net ResEnc in the proposed approach.

The cross-validation and final test results revealed a notable performance gap between the segmentation accuracy of the primary tumor (GTVp) and the nodal disease (GTVn). Although the DSC$_{agg}$ for GTVn was substantially higher than for GTVp, the HD$_{95}$ and MSD metrics were significantly larger for GTVp. This discrepancy suggests that the model struggled more with accurately delineating the nodal boundaries or even on detecting the nodes, often due to falsely predicted lymph nodes. These false predictions greatly influenced the distance based metrics.

Our experiments demonstrated that models incorporating the UMamba block achieved significant improvements in distance-based metrics (HD$_{95}$ and MSD), underscoring the value of long-range dependencies provided by the Mamba. This capability is crucial for capturing the intricate structures of HNC tumors, where understanding dependencies between primary tumors and metastatic lymph nodes is vital. Notably, the proposed UMambaAdj model, which excludes the Mamba block from the first stage, matched the ResEnc model's DSC$_{agg}$ performance while achieving HD$_{95}$ and MSD metrics similar to UMambaEnc. This suggests UMambaAdj effectively balances volumetric overlap and boundary delineation, although GTVn results indicate a need for the Mamba layer in the first stage.

The evaluation results show that while HD$_{95}$ and MSD metrics vary significantly among the methods, DSC${agg}$ remains relatively consistent. This difference is due to the metrics' sensitivities: DSC$_{agg}$, being a global overlap measure, is less affected by minor boundary discrepancies or isolated false predictions. In contrast, HD$_{95}$ is highly sensitive to boundary inaccuracies, making it more responsive to small over-segmentation, under-segmentation, or isolated false predictions. This sensitivity makes these metrics more reflective of clinically relevant errors, where even small false positives or negatives would be unacceptable. This observation underscores the importance of employing multiple evaluation metrics for a comprehensive assessment of segmentation performance. Although DSC$_{agg}$ might be suitable as a single ranking metric in public challenges due to its straightforward interpretation, relying solely on it can be misleading. The best DSC$_{agg}$ method does not always correspond to the best overall segmentation performance, particularly in accurately capturing the boundaries or avoiding false predictions—a critical factor in real-world clinical applications.

Despite the promising performance of the proposed UMambaAdj model, its effectiveness must be validated on external datasets. Our current validation was limited to a single fold from a single institutional dataset, and thus, further testing on public datasets or other private datasets is essential to confirm the generalizability and robustness of our approach. This need for broader validation is especially relevant given the rapid emergence of various Mamba-based segmentation models since Mamba's initial publication.

Recent adaptations, such as the “Swin” feature with Mamba \cite{liu2024swin}, the tri-oriented vision Mamba approach \cite{xing2024segmamba}, and the Visual Mamba U-Net \cite{wang2024mamba}, have shown that integrating Mamba with existing CNN blocks can lead to notable segmentation accuracy, all claiming achieved state-of-the-art (SOTA). However, some of these models still require significant modifications to match the 3D segmentation performance of nnU-Net. Our adjustments based on UMamba indicate that Mamba holds particular promise for HNC tumor segmentation, especially for GTVn.

Nevertheless, even with these advances, head and neck cancer tumor segmentation remains a challenging task that is far from being a "solved" problem. Fully automatic segmentation methods often face limitations that necessitate human intervention to ensure accurate treatment planning. In our study, despite achieving decent DSC$_{agg}$ scores, even SOTA models struggled with accurately identifying tumor locations, highlighting the persistent difficulties in this domain. The complex anatomy of the head and neck region, coupled with the challenge of distinguishing tumors in T2-weighted images (as evidenced in the GTVp cases from Figure 2), reinforces these challenges. Therefore, incorporating complementary information such as FDG-PET imaging \cite{jensen2021imaging,ren2021comparing}, biopsy data, patient reports \cite{rajendran2024large}, or human interaction \cite{wei2023towards} may be crucial for improving the accuracy and reliability of HNC GTV segmentation.

\section{Conclusion}
In conclusion, our customized UMambaAdj model successfully combines the strengths of long-range dependencies from UMamba blocks with the feature encoding capabilities of the nnU-Net Residual encoder, offering a balanced solution for GTV segmentation in HNC. The model showed promise in achieving accurate segmentations with a more efficient architecture, demonstrating comparable or improved performance over existing methods. However, further validation on diverse datasets and incorporating complementary tumor information with human-in-the-loop strategies will be necessary to advance the application of automatic segmentation in clinical practice for MRI guided adaptive RT.

%
%
%
%
\bibliographystyle{splncs04}  
\bibliography{reference}  
\end{document}